\documentclass[11pt,a4paper]{article}
\usepackage[hyperref]{acl2021}
\usepackage{times}
\usepackage{latexsym}

\usepackage{booktabs}
\usepackage{graphicx}
\usepackage{lipsum}
\usepackage{xspace}
\usepackage{multicol}
\usepackage{multirow}
\usepackage{mathrsfs}
\usepackage{caption}
\usepackage{amssymb}
\usepackage{enumitem}
\usepackage{changepage}
\usepackage{amsmath}
\usepackage{tabularx}
\usepackage{comment}
\usepackage{subcaption}
\usepackage{arydshln}

\usepackage{microtype}

\aclfinalcopy % Uncomment this line for the final submission
\newcommand{\ours}{\texttt{FacetSum}\xspace}

\definecolor{darkblue}{HTML}{033394}
\definecolor{darkgreen}{HTML}{005e19}
\definecolor{darkred}{HTML}{8a0000}
\definecolor{darkyellow}{HTML}{a37800}

\newcommand{\absp}{\textcolor{darkred}{Purpose}\xspace}
\newcommand{\absm}{\textcolor{darkblue}{Method}\xspace}
\newcommand{\absf}{\textcolor{darkgreen}{Findings}\xspace}
\newcommand{\absv}{\textcolor{darkyellow}{Value}\xspace}

\newcommand{\bart}{BART\xspace}
\newcommand{\bartf}{BART-Facet\xspace}

\newcommand{\code}[1]{\texttt{#1}}
\newcommand{\cmd}[1]{\textcolor{darkred}{\textbf{\small{\code{#1}}}}}
\DeclareMathOperator*{\argmax}{arg\,max}

\title{Bringing Structure into Summaries: \\ a Faceted Summarization Dataset for Long Scientific Documents}

\author{Rui Meng, Khushboo Thaker, Lei Zhang, Yue Dong, Xingdi Yuan, Tong Wang and Daqing He}

\author{Rui Meng$^\clubsuit$ \:\:\:\: Khushboo Thaker$^{\clubsuit}$ \:\:\:\: Lei Zhang$^{\clubsuit}$ \:\:\:\: Yue Dong$^{\diamondsuit}$ \:\:\:\: \\ \textbf{Xingdi Yuan$^\spadesuit$ \:\:\:\: Tong Wang$^{\spadesuit}$ \:\:\:\: Daqing He}$^{\clubsuit}$\\
$^\clubsuit$School of Computing and Information, University of Pittsburgh \\
$^\diamondsuit$ Mila / McGill University \\
$^\spadesuit$Microsoft Research, Montr\'{e}al \\
{\small \tt \{rui.meng, k.thaker, lez39, dah44\}@pitt.edu } \\
{\small \tt yue.dong2@mail.mcgill.ca } \\
{\small \tt \{eric.yuan, tong.wang\}@microsoft.com }
}
\date{}

\begin{document}
\maketitle

\begin{abstract}
Faceted summarization provides briefings of a document from different perspectives. Readers can quickly comprehend the main points of a long document with the help of a structured outline. %summarizes a document from different aspects and it can help readers quickly comprehend the content of a long text. 
However, little research has been conducted on this subject, partially due to the lack of large-scale faceted summarization datasets.
%However, faceted summarization have not drawn enough attention in prior works, partially due to the lack of large-scale annotated data to support data-driven methods.
In this study, we present \ours, a faceted summarization benchmark built on Emerald journal articles, covering a diverse range of domains.
%We present a novel dataset for this task, \ours, which consists of 60k scientific papers from Emerald and covers a wide range of domains.
Different from traditional document-summary pairs, \ours provides \textit{multiple summaries}, each targeted at specific sections
% \khu{not sure if better to say specific section or aspect of a research}
% \daqing{I think if we talk about within a paper, it is a specific section, but if we talk about within a research study, it is a specific aspect}
of a long document, including the purpose, method, findings, and value.
%Each paper in this corpus is associated with an \textit{author-provided} abstract that explicitly summarizes the \textit{purpose}, \textit{method}, \textit{findings}, and \textit{value} of their study.
Analyses and empirical results on our dataset reveal the importance of bringing structure into summaries.
%Analysis is conducted to illustrate the characteristics of structure in summaries as well as unique challenges of this dataset.
% confirms that \ours demands abilities beyond simply taking paper introduction as source text.
We believe \ours will spur further advances in summarization research and foster the development of NLP systems that can leverage the structured information in both long texts and summaries.

% Long text such as scientific publications are oftentimes structured.
% The structures can help both the authors to better organize and present their ideas, and the readers to easily retrieve and comprehend a certain part of the text (e.g., a section).
% However, the structures of text is largely neglected in prior neural summarization works, partially due to the lack of large scale labeled data that are readily accessible.
% In this work, we present a novel dataset, \ours, which consists of 60k scientific papers that cover a wide spectrum of domains.
% Each paper in \ours is associated with an \textit{author-provided} structured abstract that explicitly summarizes the \textit{purpose}, \textit{method}, \textit{findings}, and \textit{value} of their papers.
% Analysis confirms that \ours demands abilities beyond simply taking paper introduction as source text.
% We believe our corpus will spur further advances in summarization research, and foster the development of systems that can leverage the structured information in long text.

\end{abstract}

\section{Introduction}
\label{section:intro}

% what's summarization, what's structured summarization, what are previous datasets' shortcomings

% People commonly organize words following certain document or discourse structures in various genres of professional writing, such as newspaper articles~\cite{dijk1988news, yarlott2018identifying}, student essays~\cite{nguyen2018context}, corporate reports~\cite{bu2020linguistic}, legal documents~\cite{farzindar2004legal,yamada2019building}, and scientific articles~\cite{teufel2002summarizing,rimrott2007discourse,suppe1998structure}. Not only does this help authors articulate their thoughts, but also assists readers to have an overall grasp on a document. 

%\daqing{I would suggest not to put related work in appendix. If we are short of space, I suggest to add one paragraph in Introduction to be the review of the related work.}

Text summarization is the task of condensing a %usually 
long piece of text into a short summary without losing salient information.
Research has shown that a well-structured summary 
% \daqing{do not want to direct edit, so I say my comment here. What is the purpose of the phrase "any document in general" here? Can we remove it so that the meaning of the sentence is clearer: a well-structured summary can facilitate compression, which is a nice and clear statement.} 
can effectively facilitate comprehension~\cite{hartley1996obtaining,hartley1997structured}.
A case in point is the \textit{structured abstract}, which consists of multiple segments, each focusing on a specific facet of a scientific publication~\cite{hartley2014current}, such as background, method, conclusions, etc.
The structure therein can provide much additional clarity for improved comprehension and  has long been adopted by databases and publishers such as MEDLINE and Emerald.
% \daqing{since \ours uses Emerald's structured abstracts, should we say one sentence here that Emerald adopted structure abstracts in 2005?}

\begin{figure}[t!]
    \centering
    \includegraphics[width=0.5\textwidth]{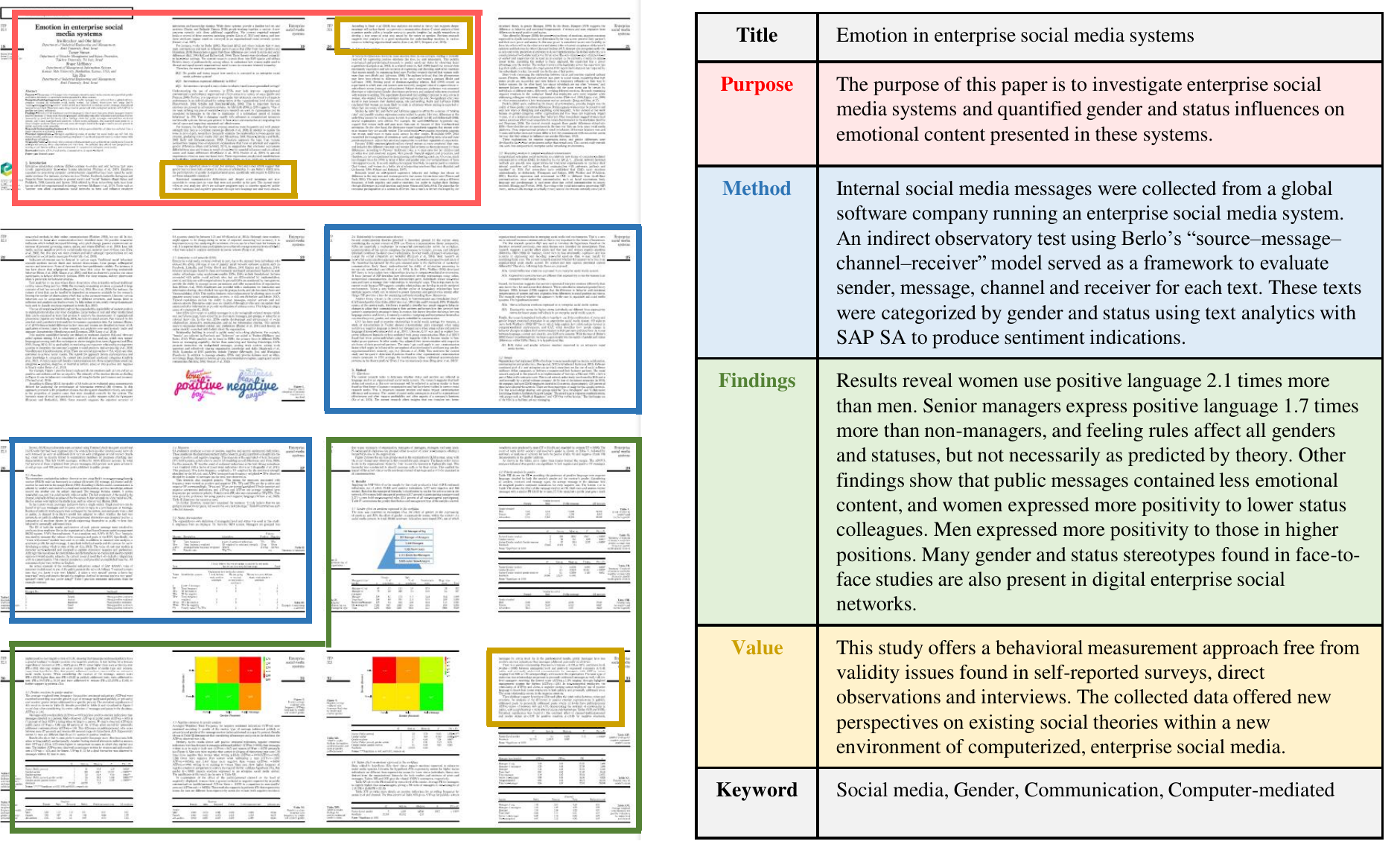}
    \caption{An example of the proposed \ours dataset. Each facet of the structured abstract summarizes different sections of the paper.
    }
    \label{fig:example}
\end{figure}
% Though some studies have explored the benefits of exploiting structures in long documents~\cite{cohan2018discourse,xu2020discourse,dong2020hiporank}, yet very few have investigated structures in summaries, largely due to the lack of large-scale data for this purpose. 

% Nevertheless, summary is oftentimes exhibited as a piece of plain text and its structure is, at large, ignored and neglected. Structured abstracts are characterised by the use of subheadings within the abstract (such as Background, Aims, Participants, Method, Results and Conclusions).

Despite these evident benefits of structure, summaries are often framed as a linear, structure-less sequence of sentences in the flourishing array of summarization studies \citep{nallapati2017summarunner,see2017get,paulus2018a,grusky2018newsroom,narayan2018don,sharma2019bigpatent,lu2020multi,cachola2020tldr}.
We postulate that a primary reason for this absence of structure lies in the lack of a high-quality, large-scale dataset with structured summaries.
In fact, existing studies in faceted summarization~\cite{huang2020coda,tauchmann2018beyond,jaidka2016overview,contractor2012using,kim2011automatic,jaidka2018insights,stead2019emerald} are often conducted with rather limited amount of data
that are  grossly insufficient to meet today's ever-growing model capacity.

We aim to address this issue by proposing the \ours dataset.
It consists of 60,024 scientific articles collected from Emerald journals,
each associated with a \emph{structured} abstract that summarizes the  article from distinct aspects including purpose, method, findings, and value.
Scale-wise, we empirically show that the dataset is sufficient for training large-scale neural generation models such as BART~\cite{lewis2020bart} for adequate generalization. %, which is in stark contrast to prior work in this area.
In terms of quality, each structured abstract in \ours is provided by the original author(s) of the article, who are arguably in the best position to summarize their own work.
We also provide quantitative analyses and baseline performances on the dataset with mainstream models in Sections~\ref{section:dataset} and 
~\ref{section:exp}.

% Besides, rather than recruiting third-party annotators to provide summaries and labels, all structured abstracts in \ours are provided by original authors, who we believe are most competent and knowledgeable for summarizing their intellectual works.
% Among a family of summarization paradigms, structured summarization is ...
% It was originally proposed by \citep{} ...
% Why previously proposed datasets do not satisfy the needs of deep neural models...

% what do we propose here
% We propose \ours, a dataset which contains XX journal papers paired with structured abstracts provided by authors... The journal papers cover a wide spectrum of topics (e.g., agriculture, business, information science)... This is not common in previous summarization datasets.

% compared to some related work, why our data is unique and necessary ...

% contribution
% The main contribution of this work is to (re)propose the task of structured summarization, and provides a large scale dataset that enables data hungry deep neural systems to learn this task...

\section{\ours for Faceted Summarization}
\label{section:dataset}

\begin{table}[t]
    % \footnotesize
    \resizebox{\columnwidth}{!}{%
    \centering
    \begin{tabular}{cccccc}
    \toprule
        \multicolumn{6}{c}{\textbf{\# documents}} \\
        \multicolumn{6}{c}{Train: 46,289 / Dev: 6,000 / Test: 6,000 / OA-Test: 2,243} \\
        \midrule
        \multicolumn{6}{c}{\textbf{\# words in abstracts}}\\
        % & Full & Purpose & Method & Findings & Value \\
        & Full & \absp & \absm & \absf & \absv \\
        mean & 290.4 & 54.1 & 52.0 & 68.6 & 47.3 \\
        std & $\pm$82.8 & $\pm$28.4 & $\pm$27.8 & $\pm$32.4 & $\pm$24.2 \\
        \midrule
        \multicolumn{6}{c}{\textbf{\# words in paper sections}}\\
        &Full & Intro. & Method & Result & Conc. \\
        recall\% & - & 84.3\% & 67.0\% & 72.4\% & 79.0\% \\
        mean & 6,827 & 885 & 1,194 & 2,371 & 747 \\
        std & $\pm$2,704 & $\pm$557 & $\pm$861 & $\pm$1,466 & $\pm$567 \\
        \bottomrule
    \end{tabular}
    }
    \caption{Statistics of the \ours dataset. }
    \label{table:stats}
\end{table}

The \ours dataset is sourced from journal articles published by Emerald Publishing\footnote{The data has been licensed to researchers at subscribing institutions to use (including data mining) for non-commercial purposes. See detailed policies at \url{https://www.emerald.com/}} (Figure~\ref{fig:example}). Unlike many publishers, Emerald imposes explicit requirements that authors summarize their work from multiple aspects~\cite{emeraldabstract}:
\textbf{\absp} describes the motivation, objective, and relevance of the research;
\textbf{\absm} enumerates specific measures taken to reach the objective, such as experiment design, tools, methods, protocols, and datasets used in the study;
\textbf{\absf} present major results such as answers to the research questions and confirmation of hypotheses;
and \textbf{\absv} highlights the work's value and originality\footnote{There are three optional facets (about research, practical and social implications) that are missing from a large number of articles and hence omitted in this study.}.
Together, these facets give rise to a comprehensive and informative structure in the abstracts of the Emerald articles,
and by extension, to \ours's unique ability to support faceted summarization.

\subsection{General Statistics}
\label{sec:general_stats}
We collect 60,532 publications from Emerald Publishing spanning 25 domains.
% We ensure the distribution over research domains to be consistent across the three sets.
%For each set, we ensure papers from each research domain are divided evenly.
Table~\ref{table:stats} lists some descriptive statistics of the dataset.
Since \ours is sourced from journal articles, texts therein are naturally expected to be longer compared to other formats of scientific publications.
In addition, although each facet is more succinct than the traditional, structure-less abstracts, a full length abstract containing all facets can be considerably longer.
Empirically, we compare the source and the target lengths 
% \daqing{source and target is probably clear to a person working on summarization, but it is not easy to know that source means the full document and target means the abstract.}
with some existing summarization datasets in similar domains including CLPubSum \citep{collins2017supervised}, PubMed \citep{cohan2018discourse}, ArXiv \citep{cohan2018discourse}, SciSummNet \citep{yasunaga2019scisummnet}, and SciTldr \citep{cachola2020tldr}.
On average, the source length in \ours is 58.9\% longer (6,827 vs 4,297), and the target length is 37.0\% longer (290.4 vs 212.0).

From a summarization perspective, these differences imply that \ours may pose significantly increased  modeling and computation challenges due to the increased lengths in both the source and the target.
Moreover, the wide range of research domains (Figure~\ref{fig:category}, Appendix~\ref{appendix:domain}) may also introduce much linguistic diversity w.r.t. vocabulary, style, and discourse. Therefore, compared to existing scientific publication datasets that only focus on specific academic disciplines~\cite{cohan2018discourse,cachola2020tldr}, \ours can also be used to assess a model's robustness in domain shift and systematic generalization.

% Moreover, the wide range of research domains covered by \ours 
% presents a great diversity in language use. Both challenges are unique and (Figure~\ref{fig:category}) in stark contrast to existing scientific publication datasets~\cite{cohan2018discourse,cachola2020tldr} that usually focus on highly specific academic disciplines. 

% \daqing{should remove "by \ours", also what is this unique challenge? It is not clear here}

% collins2017supervised
% yasunaga2019scisummnet
% lev2019talksumm
% cachola2020tldr
% cohan2018discourse

% 1. journal paper is longer than arxiv/pubmed/tldr?
% 2. structured abstract is longer than traditional ones?

To facilitate assessment of generalization, we reserve a dev and a test set each consisting of 6,000 randomly sampled data points; the remaining data are intended as the training set.
We ensure that the domain distribution is consistent across all three subsets. Besides, we intentionally leave out Open-Access papers as another test set, to facilitate researchers who do not have full Emerald access\footnote{
% Precise split information will be released together with the dataset for reproducibility. 
Both the split information of \ours and the code for scraping and parsing the data are available at \url{https://github.com/hfthair/emerald_crawler}}. 

\begin{figure*}[t!]
    \centering
    \includegraphics[width=1.0\textwidth]{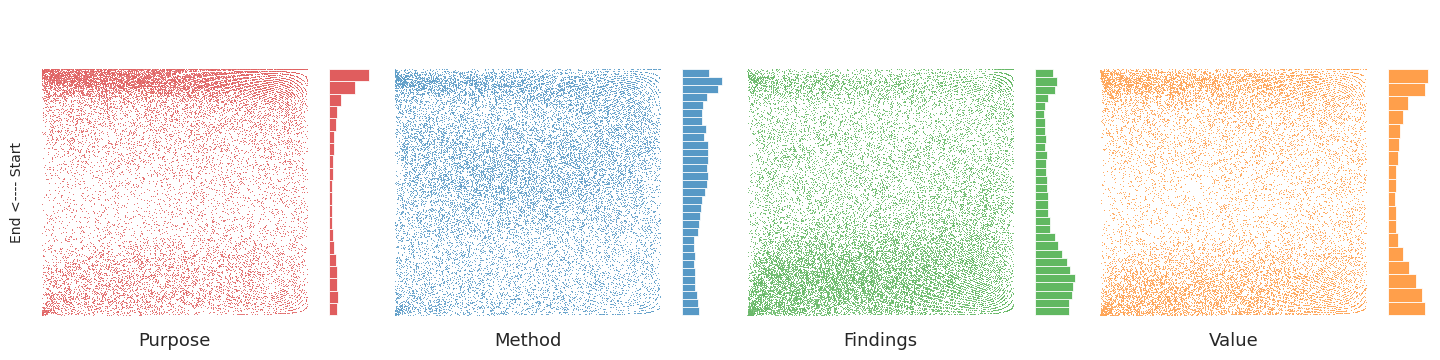}
    \caption{Oracle sentence distribution over a paper. X-axis: 10,000 papers sampled from \ours, sorted by full text length from long to short; y-axis: normalized position in a paper. We provide each sub-figure's density histogram on their right.}
    \label{fig:distribution}
\end{figure*}

\subsection{Structural Alignment}
\label{subsection:structural_alignment}
% \subsection{Positional Alignment}
% \#\#\# DRAFT BEGINS \#\#\#
% \subsubsection*{Motivations and Notations}
In this section, we focus our analysis on one of the defining features of \ours\ --- its potential to support faceted summarization.
% Specifically, we investigate the alignment between the articles and their corresponding abstracts (as summarization sources and targets),
% with a particular interest in
Specifically, we investigate how the abstract structure (i.e., facets) aligns with the article structure.
Given an abstract facet $A$ and its corresponding article $S$, we quantify this alignment by:
\begin{equation}
    \label{eqn:s_a_alignment}
    \small
    S_A=\{\argmax_{s_i\in S}(\texttt{Rouge-1}(s_i,a_j)):a_j\in A\}
\end{equation}
Semantically, $S_{A}$ consists of sentence indices in $S$ that best align with each sentence in $A$.

% Formally, given sentence $s$ and a set of sentences $A$, $f(s,A)$ denotes the Rouge-1 score between $s$ and the concatenation of all sentences in $A$.
% We can then compute a ranking $\{\rho_{1:n}\}$ on a set of $n$ sentences $S=\{s_{1:n}\}$ \emph{conditioned on} $A$, such that, $f(s_{\rho_i},A)>f(s_{\rho_{i+1}},A),\forall i=1:n-1$.

% Finally, we denote $S_{A}=\{s_{\rho_i}:i=1:|A|\}$, where $|A|$ is the number of sentences in $A$.
% When $S$ is an article an abstract $A$, $S_{A}$ is essentially an extractive summary of $S$ based on both the content and the length of its abstract. 

%\subsubsection*{Sentence-level Alignment}
\noindent\textbf{Sentence-level Alignment}\quad
We first plot the tuples $\{(s_i,i/|S|):i\in S_{A}\}$, where $s_i$ is the $i$-th sentence in $S$, and $|S|$ is the number of sentences in $S$.
Intuitively, the plot density around position $i/|S|$ entails the degree of alignment between the facet $A$ and the article $S$ at that position\footnote{We use the relative position $i/|S|$ so that all positions are commensurate across multiple documents.}.
With 10,000 articles randomly sampled from \ours, Figure~\ref{fig:distribution} exhibits distinct differences in the density distribution among the facets in \ours.
For example, with $A=$ \absp, resemblance is clearly skewed towards the beginning of the articles,
while \absf are mostly positioned towards the end; the \absm distribution is noticeably more uniform than the others.
These patterns align well with intuition, and are further exemplified by the accompanying density histograms.

\begin{table}[t!]
    \centering
    \includegraphics[width=0.45\textwidth]{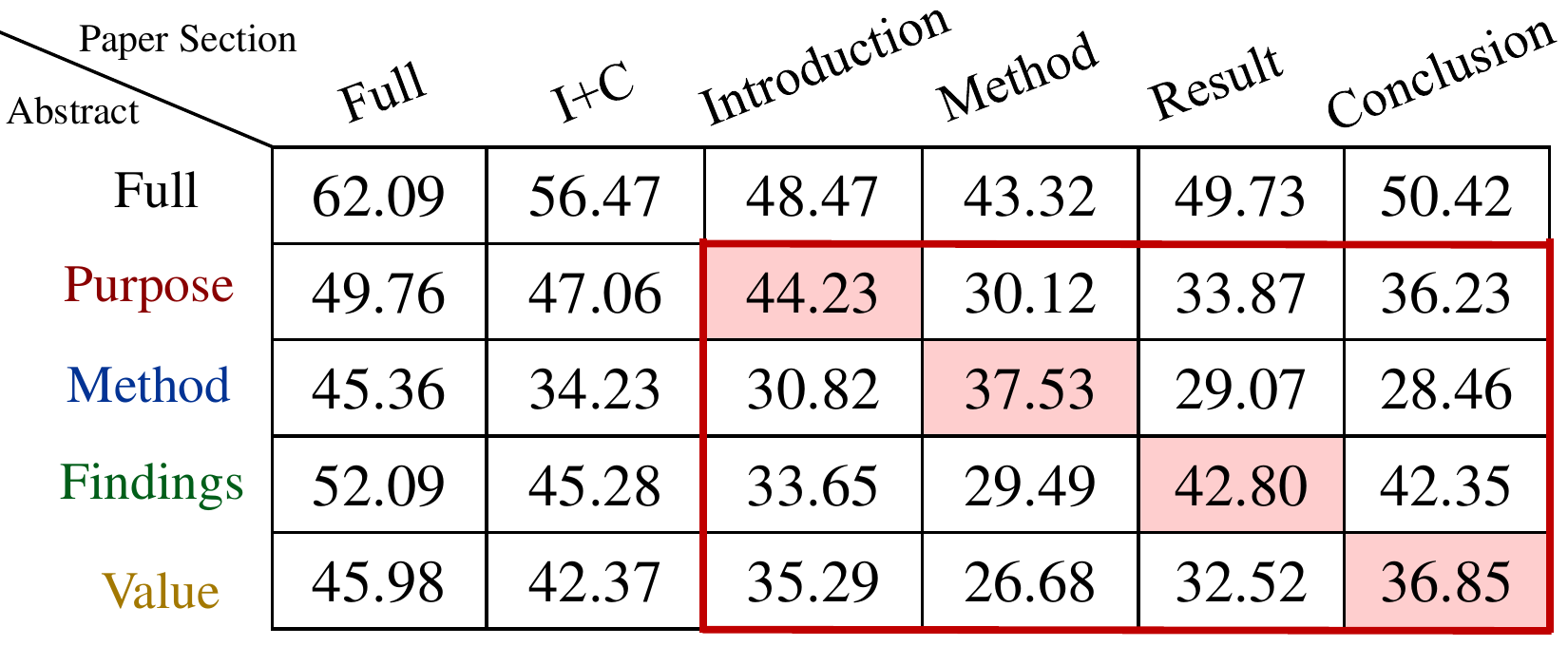}
    \caption{Scores of sentence aligning in Rouge-L.}
    \label{table:oracle_scores_rl}
\end{table}

% \subsubsection*{Section-level Alignment}
%\paragraph{Section-level Alignment}
\noindent\textbf{Section-level Alignment}\quad
We now demonstrate how different abstract facets align with different sections in an article.
Following conventional structure of scientific publications~\cite{suppe1998structure,rosenfeldt2000write}, we first classify sections into \textit{Introduction}, \textit{Method}, \textit{Result} and \textit{Conclusion} using keyword matching in the section titles.\footnote{
To ensure close-to-perfect precision, we choose keywords that are as specific and prototypical to each section as possible (listed in Appendix~\ref{sec:cue_word}).
The resulting recall is around $0.7$, i.e. about 70\% of sections can be correctly retrieved with the title-keyword matching method. And we find 2,751 (out of 6,000) test samples that all four sections are matched successfully.
Though far from perfect, we believe this size is sufficient for the significance of subsequent analyses.}
% If a section fails to match any keywords, we take the full text instead.

Given a section $S^i\subseteq S$ and an abstract $A_j\subseteq A$, we define the section-level alignment $g(S^i,A_j)$ as $\texttt{Rouge-1}(\texttt{cat}(S^i_{A_j}),\texttt{cat}(A_j))$,
where $\texttt{cat}(\cdot)$ denotes sentences concatenation, and $S^i_{A_j}$ is defined by Equation~\eqref{eqn:s_a_alignment}.
Table~\ref{table:oracle_scores_rl} is populated by varying $A_j$ and $S^i$ across the rows and columns, respectively.
\textbf{Full} denotes the full paper or abstract (concatenation of all facets). 
We also include the concatenation of introduction and conclusion (denoted I+C) as a possible value for $S^i$, due to its demonstrated effectiveness as summaries in prior work \cite{cachola2020tldr}.

The larger numbers on the diagonal (in red) empirically confirm a strong alignment between \ours facets and their sectional counterparts in articles.
We also observe a significant performance gap between using I+C and the full paper as $S^i$.
% This is exciting because in prior scientific document summarization works, I+C is often used as a standard workaround to tackle long source text \eric{cite someone}. Partially due to the lack of details in their ground-truth summaries, using I+C is shown to contain sufficient information.
One possible reason is that the summaries in \ours (particularly \absm and \absf) may contain more detailed information beyond introduction and conclusion.
This suggests that for some facets in \ours, simple tricks to condense full articles do not always work; models need to instead comprehend and retrieve relevant texts from full articles in a more sophisticated manner.

\begin{table*}[t!]
    % \fontsize{8}{8}\selectfont 
    % \scriptsize
    \resizebox{\linewidth}{!}{%
    \centering
    \begin{tabular}{clcccccc}
    \toprule\midrule
        & \textbf{Model} & \textbf{Source Text} &  \textbf{Full} & \textbf{\absp} & \textbf{\absm} & \textbf{\absf} & \textbf{\absv} \\ 
        \midrule
        \midrule
        \multicolumn{8}{l}{\ours \texttt{Test}} \\ 
        \midrule
        Oracle & Greedy Extractive~\cite{nallapati2017summarunner} & corresponding & 60.39 & 44.66 & 41.00 & 46.44 & 38.10 \\ 
        \midrule
        %\multicolumn{6}{c}{\scriptsize{Unsupervised Models}}\\
        %\midrule
        Heuristic & Lead-K & corresponding               & 36.78 & 17.83 & 15.29 & 15.92 & 16.08 \\ 
        Models & Tail-K &    sections                    & 33.31 & 21.67 & 12.62 & 16.66 & 17.43 \\ 
        \midrule
        & SumBasic \citep{vanderwende2007beyond} &          & 38.71 & 18.17 & 15.41 & 16.31 & 16.57 \\ 
        Unsupervised& LexRank \citep{erkan2004lexrank} &  corresponding               & 42.18 & 18.72 & 16.23 & 18.11 & 17.75 \\ 
        Models& LSA \citep{gong2001lsa} &  sections                         & 35.98 & 18.29 & 15.86 & 16.92 & 16.62 \\ 
        & TextRank \citep{mihalcea2004textrank} &                      & 41.87 & 21.67 & 13.62 & 18.63 & 19.23 \\ 
        & HipoRank \citep{dong2020hiporank} &                      & 42.89 & 22.73 & 15.20 & 18.38 & 19.68 \\ 
        \midrule
        %\multicolumn{6}{c}{\scriptsize{Abstractive Models}}\\
        %\midrule
        & \bart \citep{lewis2020bart} &  I+C                     & 44.36 & 41.14 & 20.75 & 14.72 & 5.85 \\
        Supervised & \bartf  & I+C               & \textbf{47.09} & \textbf{43.47} & \textbf{29.07} & \textbf{30.97} & \textbf{28.90} \\
        Models & \bart  & full paper    & 42.74 & 41.21 & 20.53 & 14.33 & 5.07 \\
        & \bartf & full paper                & 45.76 & 42.55 & 28.07 & 28.98 & 28.70 \\
        \midrule
        \midrule
        \multicolumn{8}{l}{\ours \texttt{OA-Test}} \\ 
        \midrule
        & \bart  &  I+C                     & 44.97 & 43.51 & 26.73 & 11.79 & 0.31 \\
         & \bartf  & I+C                
        & 51.32 & 43.66 & 30.16 & 32.22 & 29.68 \\       
        % \midrule
        % \midrule
        % \multicolumn{8}{l}{\ours \texttt{Open-Access train-100k}} \\ 
        % \midrule
        % \midrule
        %     & \bart 
        %     &  I+C                     & 45.96 & 42.46 & 23.84 & 11.62 & 0.72 \\
        % Supervised & \bartf  & I+C                & 49.06 & 44.75 & 30.67 & 32.57 & 29.81 \\       
        \bottomrule
    \end{tabular}}
    \caption{Model performance on \ours (Rouge-L). See Table~\ref{table:oracle_scores_full}  and \ref{table:model_performance_full} in Appendix~\ref{appendix:full_results} for full results. \textbf{Bold} text indicates the best scores on \ours test split in each column.}
    
    \label{table:model_performance_rl}
\end{table*}

\section{Experiments and Results}
\label{section:exp}
%on it with 
We use \ours to benchmark a variety of summarization models from state-of-the-art supervised models to unsupervised and heuristics-based models.
We also provide the scores of a sentence-level extractive oracle system~\cite{nallapati2017summarunner}. We report Rouge-L in this section and include Rouge-1/2 results in Appendix~\ref{appendix:full_results}.

%\paragraph{Unsupervised Models vs Heuristics}
\noindent\textbf{Unsupervised Models vs Heuristics}\quad
%\label{subsection:unsupervised_vs_heuristics}
%We first compare the performances of heuristics and unsupervised systems on \ours.
We report performances of unsupervised and heuristics summarization methods  (see Table~\ref{table:model_performance_rl}). 
Tailoring to the unique task of generating summaries for a specific facet, we only use the section (defined in Section~\ref{subsection:structural_alignment}) corresponding to a facet as model input.
Evaluation is also performed on the concatenation of all facets (column \textbf{Full}), which resembles the traditional research abstract.
% Two simple heuristics are designed based on the idea that sentences at the beginning/end tend to be high-level synopses of the text:
\textbf{Lead-K}/\textbf{Tail-K} are two heuristic-based models that extract the first/last $k$ sentences from the source text.% where $k$ is the number of sentences in ground-truth.

%including SumBasic, LexRank, LSA and TextRank.

We observe that heuristic models do not perform well on \textbf{Full}, where the unsupervised models can achieve decent performance.
Nevertheless, all models perform poorly on summarizing individual facets, and unsupervised models fail to perform better than simple heuristics consistently.
% , even in the cases where the matching paper sections are provided as source text.
The inductive biases of those models may not be good indicators of summary sentences on specific facets. A possible reason is that they are good at locating overall important sentences of a document, but they cannot differentiate sentences of each facet, even we try to alleviate this by using the corresponding section as input. %Developing methods to direct their outputs in a controlled way would be an interesting direction for future work.
% \eric{What are some other observations, what can we conclude from this?}

% \eric{Compare unsupervised systems with heuristics, show that unsupervised models are not significantly working better than simple heuristics --- and thus \ours is not for unsupervised models.}

%\paragraph{Supervised Models}
\noindent\textbf{Supervised Models}\quad
%\label{subsection:supervised}
As for the supervised baseline, we adopt the \bart model~\citep{lewis2020bart}, which has recently achieved SOTA performance on abstractive summarization tasks with scientific articles \citep{cachola2020tldr}.
We propose two training strategies for the \bart model, adapting it to handle the unique challenge of faceted summarization in \ours.
In \textbf{\bart}, we train the model to generate the concatenation of all facets, joined by special tokens that indicate the start of a specific facet (e.g., \cmd{|PURPOSE|} to indicate the start of \absp summary). 
During evaluation, the generated text is split into multiple facets based on the special tokens, and each facet is compared against the corresponding ground-truth summary.
In \textbf{\bartf}, we train the model to generate one specific facet given the source text and an indicator specifies which facet to generate.
Inspired by CATTS \citep{cachola2020tldr}, we prepend section tags at the beginning of each training input to generate summaries for a particular facet (see implementation details in Appendix \ref{ref:bart_implementation}). 

Empirically, supervised models outperform unsupervised baselines by a large margin (Table~\ref{table:model_performance_rl}).
Comparing between the two training strategies, \bartf  outperforms \bart significantly.
While \bart performs comparably on \absp, performance decreases drastically for subsequent facets, possibly due to current models' inadequacy with long targets.
%As shown in Table ~\ref{table:stats}, the full abstract contains 290 tokens on average (possibly more if using sub-word) as opposed to 60 tokens in single facet.
Thus it can perform decently at the beginning of generation ($\approx$40 on \absp), where the dependency is relatively easy-to-handle. 
However, the output quality degrades quickly towards the end ($\approx$5 on \absv).

% It is also obvious that both the two training strategies perform better using I+C as source text regardless the oracle system performs significantly better with full paper than I+C (Table~\ref{table:oracle_scores_rl}).
With I+C as source text, both training strategies exhibit much better results than using full paper. This is opposite to the observation in Table~\ref{table:oracle_scores_rl}, potentially due to the limitation of the current NLG systems, i.e., the length of source text has crucial impacts to the model performance.
With the much extended positional embeddings in our models (10,000 tokens), we suspect some other issues such as long term dependencies %are causing 
may lead to this discrepancy, which warrants further investigation.

\section{Related Work}
We acknowledge several previous efforts towards faceted summarization. Prior to our study, generating structured abstracts for scientific publications was also discussed in~\cite{gidiotis2019structured, gidiotis2020divide}. The authors built a structured abstract dataset PMC-SA, consisting of 712,911 biomedical articles from PubMed Central, and they proposed to summarize a paper from a specific facet by taking corresponding sections as inputs. Compared with their works, \ours covers a wider range of academic fields and we provide in-depth discussions on the structured abstract to justify its value as a novel NLP challenge.
Our research shares some resemblance to studies on abstract sentence classification, whose goal is to classify abstract sentences into several facets, instead of summarizing the full text. MEDLINE is commonly used for this task~\cite{kim2011automatic}, so as the Emerald data~\cite{stead2019emerald}. A recent study~\cite{huang2020coda} introduced a new dataset CODA, in which 10,966 abstracts are split into subsentences and labelled into five categories by third-party annotators. However, we think scientific documents are generally difficult to comprehend for people without specific training, thus original authors are in the best position to summarize their own work. 
Faceted summarization was also involved in CL-SciSumm 2016 Shared Task~\cite{jaidka2018insights}, where the faceted summary of a paper is defined as the citation sentences in its successor studies, since new studies typically describe previous work from different perspectives. However, this idea may not easily scale up in the real world since many papers do not have enough number of citations, especially for newly published ones.

\section{Conclusion \& Future Work}
\label{section:discussion}
% Comparison to related works
We introduce  \ours to support the research of faceted summarization, which targets summarizing scientific documents from multiple facets. We provide extensive analyses and results to investigate the characteristics of \ours.  Our observations call for the development of models capable of handling very long documents and outputting controlled text. Specifically, we will consider exploring the following topics in future work: (1) incorporating methods for long-document processing, such as reducing input length by extracting key sentences~\citep{pilault2020extractive} or segments~\cite{zhao2020seal}; (2) examining the possibility of building a benchmark for systematic generalization~\cite{bahdanau2018systematic} with the domain categories; (3) automatically structuring traditional abstracts~\cite{huang2020coda} with \ours.
% 1. explore possibility of benchmark for systematic generalization; 2. convert traditional abstracts to structured ones; 3. apply to discourse/document structure analysis in scientific documents~\cite{huang2020coda}; 4. explore other genres of corpora.

% \clearpage
\bibliographystyle{acl_natbib}
%\bibliography{anthology,acl2021}
\bibliography{acl2021}

%\appendix
\clearpage
\appendix
\section{Keyword List for Identifying Paper Sections}
\label{sec:cue_word}
\begin{table}[!ht]
    \centering
    \fontsize{10}{10}\selectfont 
    \begin{tabular}{ll}
        \toprule\midrule
        \textbf{Category} & \textbf{Keyword} \\ \midrule
        Introduction & intro, purpose \\
        Method & design, method, approach \\
        Result & result, find, discuss, analy \\
        Conclusion & conclu, future \\
        \bottomrule
    \end{tabular}
    \caption{Keywords for identifying paper sections used in Section~\ref{subsection:structural_alignment}.}
    \label{table:cue_word}
\end{table}

\section{Most Frequent Words in Each Abstract Facet}
\label{sec:top_word_each_facet}
\begin{table}[!ht]
    \centering
    \fontsize{10}{10}\selectfont 
    \begin{tabular}{c  p{0.2\linewidth}  p{0.2\linewidth}  p{0.2\linewidth}}
        \toprule\midrule
        \textbf{Facet} & \textbf{Verb}  & \textbf{Noun}  & \textbf{Adjective} \\ 
        \midrule
        \absp 
        & aim  &   paper   &   social     \\
        & examin   &   purpos   &  new     \\
        & investig &	studi &	organiz	  \\
        & explor &	manag &	differ		  \\
        & develop &      research &    public \\
        \midrule 
        \absm 
        & base &         studi &  structur	     \\
        & conduct &          data &    qualit	 \\
        & collect &       analysi &    differ	  \\
        & test &         model &     empir	  \\
        & develop &         paper &    social  \\
        \midrule
        \absf
        & found &        result &  signific     \\
        & indic &         studi &     posit     \\
        & suggest &         manag &    social	 \\
        & provid &        effect &    differ	 \\
        & identifi &  relationship &    higher	 \\
        \midrule
        \absv
        & provid &         studi &       new    \\
        & contribut &         paper &    social  \\
        & develop &      research &    differ  \\
        & base &         manag &     empir	  \\
        & examin &     literatur &    import  \\
        \bottomrule
    \end{tabular}
    \caption{Top five frequent verbs/nouns/adjectives in each facet of structured abstract. We preprocess the text with lowercasing, stemming and stopword removal and extract part-of-speech tags using Spacy~\cite{spacy}.}
\end{table}

\section{Implementation Details}
\label{ref:bart_implementation}
To make BART take full text as input, we extend the positional embedding to 10,000 tokens. This was required to leverage long text of papers in \ours with average length of 6000 words. 

Experiments of unsupervised baselines are implemented with Sumy~\cite{sumy} and official code of HipoRank. We tune the hyperparameters of HipoRank with the validation set. The BART experiments are finetuned using Fairseq~\cite{ott2019fairseq}, with learning rate of $3e^{-5}$, batch size of 1, max tokens per batch of 10,000 and update frequency of 4. We finetune all models for  20,000 steps with single NVIDIA Tesla V100 16GB and we report the results of the last checkpoint. The small batch size is the consequence of the large input size. For inference, we use beam size of 4 and maximum length of 500/200 tokens for \bart/\bartf respectively.

\section{Domains Covered by \ours}
\label{appendix:domain}
\begin{figure*}[ht!]
    \centering
    \includegraphics[width=0.8\textwidth]{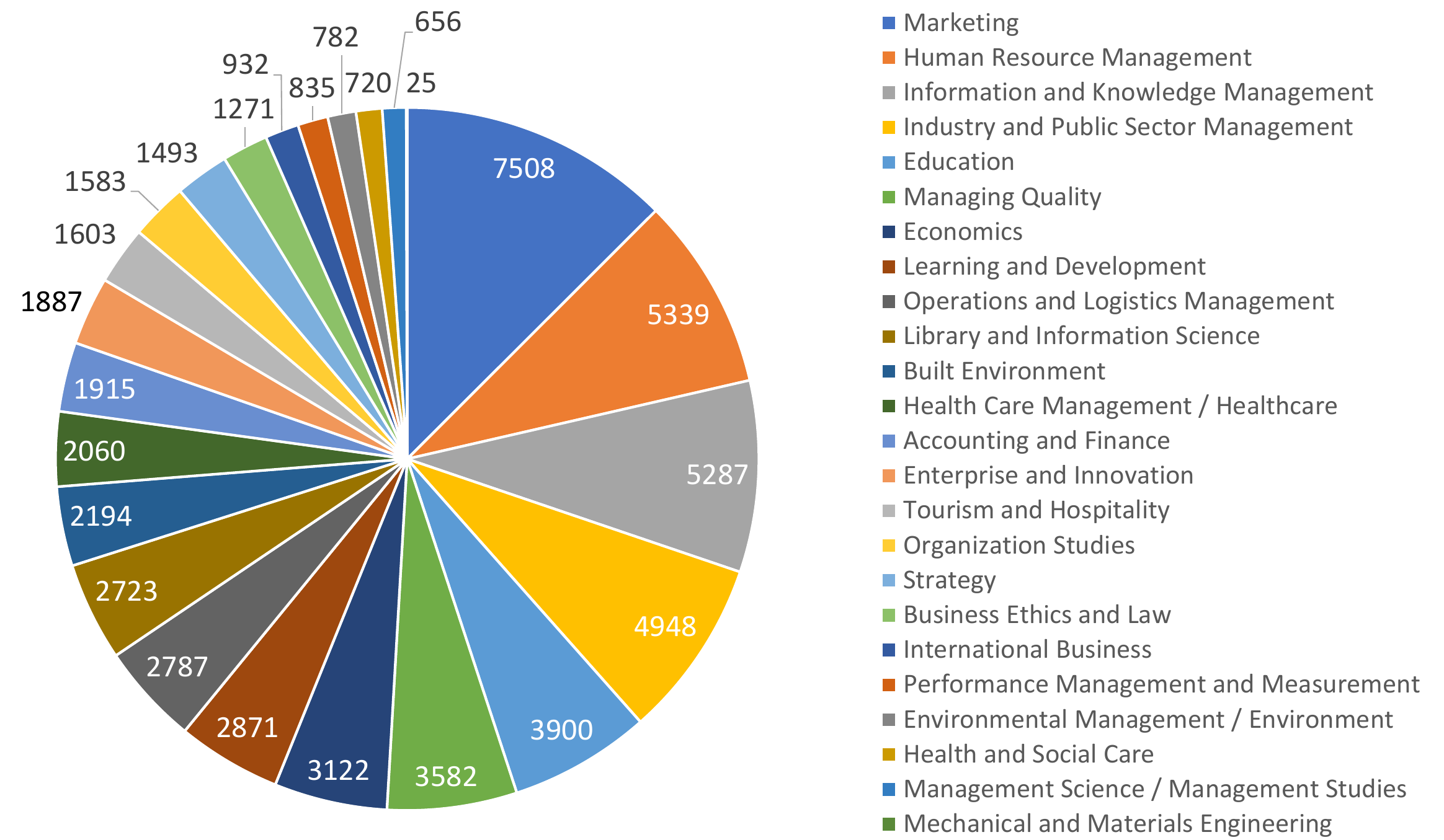}
    \caption{Data distribution of domain categories, sorted in descending order.}
    \label{fig:category}
\end{figure*}

In Figure~\ref{fig:category}, we show the distribution of domain categories in \ours.

\section{Full Results}
\label{appendix:full_results}
In this section, we provide additional experiment results.
In Table~\ref{table:oracle_scores_full}, we show the full results of the extractive oracle system (first row in Table~\ref{table:model_performance_rl}).
In Table~\ref{table:model_performance_full}, we provide full results of all other models (heuristic models, unsupervised models, and supervised models in Table~\ref{table:model_performance_rl}).

\begin{table*}
    \centering
    \resizebox{\linewidth}{!}{
    \fontsize{10}{10}\selectfont 
    \renewcommand{\arraystretch}{1.35}
    \begin{tabular}{lccccc}
    \toprule\midrule
        \textbf{R1/R2/RL} & \textbf{Full} & \textbf{\absp} & \textbf{\absm} & \textbf{\absf} & \textbf{\absv} \\ \midrule
        Full$_{body}$ & 64.92/33.75/60.39 & 57.35/30.24/49.42 & 53.30/26.40/45.58 & 59.30/33.25/52.42 & 53.39/26.84/45.55 \\ 
        IC$_{body}$ & 58.82/28.42/54.17 & 53.60/27.13/45.73 & 43.13/17.08/35.64 & 52.03/25.90/44.86 & 48.97/22.84/41.09 \\
        \midrule
        Intro$_{body}$ & 53.32/22.96/48.59 & \textbf{52.51/26.48/44.66} & 41.27/16.05/34.03 & 44.67/17.49/37.10 & 44.65/17.80/36.47 \\ 
        Method$_{body}$ & 52.05/20.52/47.35 & 45.16/16.61/36.84 & \textbf{48.60/21.67/41.00} & 44.77/17.69/37.67 & 40.94/13.55/32.94 \\ 
        Result$_{body}$ & \textbf{56.85/23.79/51.97} & 47.90/18.07/38.96 & 42.31/14.46/34.41 & \textbf{53.71/26.32/46.44} & 44.93/16.91/36.66 \\ 
        Conclu$_{body}$ & 55.26/25.26/50.58 & 47.76/18.88/38.94 & 40.53/13.84/32.83 & 51.81/25.81/44.73 & \textbf{46.14/19.66/38.10} \\ \bottomrule
    \end{tabular}}
    \caption{Full results (Rouge-1/2/L) of the extractive oracle system~\cite{nallapati2017summarunner} on \ours. \textbf{Bold} text indicates the best scores in the lower four rows in each column.}
    \label{table:oracle_scores_full}
\end{table*}

\begin{table*}
    \centering
    \fontsize{8.2}{10}\selectfont 
    \renewcommand{\arraystretch}{1.25}
    \begin{tabular}{lccccc}
    \toprule\midrule
        \textbf{R1/R2/RL} & \textbf{Full} & \textbf{\absp} & \textbf{\absm} & \textbf{\absf} & \textbf{\absv} \\ 
        \midrule\midrule
        \multicolumn{6}{l}{\ours \texttt{Test}} \\ 
        \midrule
        Lead-K & 39.65/11.01/36.78 & 21.95/4.89/17.83 & 18.69/5.94/15.29 & 18.84/4.31/15.92 & 20.14/3.05/16.08 \\ 
        Tail-K & 35.90/10.96/33.31 & 25.48/7.23/21.67 & 14.88/2.64/12.62 & 19.25/4.41/16.66 & 20.90/4.71/17.43 \\ \midrule
        SumBasic & 42.11/10.01/38.71 & 22.23/4.68/18.17 & 18.40/5.02/15.41 & 19.15/3.93/16.31 & 20.64/3.08/16.57 \\ 
        LexRank & 46.35/15.12/42.18 & 22.97/5.28/18.72 & 19.44/5.84/16.23 & 21.66/5.66/18.11 & 22.39/4.05/17.75 \\ 
        LSA & 39.84/9.59/35.98 & 22.47/4.91/18.29 & 19.10/5.58/15.86 & 20.29/4.59/16.92 & 20.96/3.31/16.62 \\ 
        TextRank & 46.90/16.04/41.87 & 28.29/9.39/21.67 & 17.55/4.32/13.62 & 23.90/7.17/18.63 & 25.99/7.07/19.23 \\ 
        HipoRank & 46.48/15.42/42.89 & 27.71/8.29/22.73 & 18.27/4.65/15.20 & 21.75/5.31/18.38 & 24.54/5.26/19.68 \\ \midrule
        BART I+C & 47.21/19.59/44.36 & 46.61/27.10/41.14 & 23.85/7.98/20.75 & 16.84/5.34/14.72 & 7.21/1.93/5.85 \\ 
        % \midrule
        BART-Facet I+C & \textbf{50.62/20.97/47.09} & \textbf{49.59/28.70/43.47} & \textbf{34.61/11.82/29.07} & \textbf{36.42/12.63/30.97} & \textbf{35.37/11.75/28.90} \\ 
        % \midrule
        BART full body & 45.49/18.10/42.74 & 46.74/27.09/41.21 & 23.66/7.92/20.53 & 16.39/4.63/14.33 & 6.30/1.62/5.07 \\ 
        % \midrule
        BART-Facet full body & 49.29/19.60/45.76 & 48.65/27.72/42.55 & 33.49/11.01/28.07 & 34.46/10.49/28.98 & 35.27/11.44/28.70 \\ 
        \midrule
        \midrule
        \multicolumn{6}{l}{\ours \texttt{OA-Test}} \\ 
        \midrule
        % \midrule
        % \midrule
        BART I+C & 
        48.85/20.84/44.97 & 49.43/29.44/43.51 & 31.1/10.16/26.73 & 13.78/4.45/11.79 & 0.4/0.1/0.31 \\
        % \midrule
        BART-Facet I+C & 
        48.31/22.63/51.32 & 49.59/28.69/43.66 & 35.82/12.84/30.16 & 37.46/14.02/32.22 & 35.9/12.75/29.68 \\
        % \midrule
        % \midrule
        % \multicolumn{6}{l}{\ours \texttt{OA-Test Train-100k}} \\ 
        % \midrule
        % \midrule
        % BART I+C\_1 & 49.92/21.28/45.95 & 47.99/28.45/42.46 & 27.55/9.31/23.84 & 13.59/4.49/11.62 & 
        % 0.84/0.32/0.72 \\
        % % \midrule
        % BART-Facet I+C & 52.31/23.06/49.06 & 51.09/30.55/44.75 & 36.52/13.43/30.66 & 38.23/14.36/32.58 & 36.28/12.79/29.81 \\
        \bottomrule
    \end{tabular}
    \caption{Full results (Rouge-1/2/L) of different models on \ours. \textbf{Bold} text indicates the best scores on \ours test split in each column.}
    \label{table:model_performance_full}
\end{table*}

\section{Example of Outputs by \bart and \bartf}
\label{sec:example_generation}
In Table~\ref{table:example_generation}, we show an example of the generated faceted summaries by \bart and \bartf of the same paper, compared against the ground-truth faceted abstract.

\begin{table*}[!ht]
    \centering
    \resizebox{\linewidth}{!}{
    \fontsize{10}{10}\selectfont 
    \begin{tabular}{m{0.07\linewidth}  p{0.3\linewidth}  p{0.3\linewidth}  p{0.3\linewidth}}
        \toprule\midrule
        \textbf{Facet} &\textbf{ Ground Truth} &\textbf{  \bart} & \textbf{\bartf} \\ 
        \midrule
        \textbf{\absp} &
        The purpose of this paper is to advance the information society research by examining and better understanding the impact of the adoption information and communication technologies (ICT) within households on improving sustainability.
        & 
        The purpose of this paper is to answer the question: Does the adoption of information and communication technologies (ICT) within households contribute to sustainability? For this purpose, the study focuses on addressing the following specific questions: What criteria clearly and deeply describe the ICT adoption within households? What are the criteria that clearly describe sustainability within households, and are there statistically significant correlations between ICT and sustainability?
        &
The purpose of this paper is to answer the question: Does the adoption of information and communication technologies (ICT) within households contribute to sustainability?        \\
        \midrule 
        \textbf{\absm} &
        The survey questionnaire was used and data collected from 679 Polish households were statistically analyzed to understand the phenomena of ICT adoption and sustainability as well as identify correlations between them. &
A survey was carried out in Poland. The data were collected through a structured questionnaire and analyzed using structural equation modeling (SEM).
        &
The study is based on a quantitative research design. The data were collected through a questionnaire survey among the households in Poland. The survey was carried out in two phases. In the first phase, the ICT adoption within households was measured using a structured questionnaire, and in the second phase the data were analyzed using structural equation modeling (SEM) to test the hypothesized relationships.
        \\ \midrule
        \textbf{\absf} &
        The research findings reveal that the ICT adoption is well described by the ICT outlay, information culture, ICT management and ICT quality, whereas sustainability is composed of ecological, economic, socio-cultural and political sustainability. Furthermore, the ICT quality, ICT management and information culture have a significant impact on sustainability, whereas the ICT outlay does not have such an impact. &
        Results show that ICT is a key enabler of sustainability in households. The results also show that there are statistically significant correlation between the IIT adoption within the households and sustainability.

&        
The results show that the adoption of ICT within households is positively related to sustainability. The results also show that there are statistically significant correlations between the ICT adoption within households and sustainability.

 \\ \midrule
         
 \textbf{\absv} & 
 The paper provides and verifies a new theoretical model of sustainable information society to depict various dimensions shaping the ICT adoption and their impact on different types of sustainability in the context of households. &
 This study is the first to empirically investigate the impact of ICT on sustainability. The findings of this study will be complementary with findings concerning the contribution of IIT to sustainability in enterprises and allow for the advancement in the sustainable information society (SIS) research.
&
 This study contributes to the literature by providing a deeper understanding of the ICT adoption within households and the contribution of ICT to sustainability in transition economies, i.e. the former European Eastern Bloc countries.
\\
        \bottomrule
    \end{tabular}}
    \caption{Outputs by \textbf{\bart} and \textbf{\bartf} on different facets. Both models are able to generate reasonable summaries given the specified facet. \bartf provides more information of \absm and less errors than \bart (e.g. ``IIT'' is a typo of ``ICT''). However both models tend to directly copy text from the source, for example both outputs of \absp can be found in the introduction of the paper.}
    \label{table:example_generation}
\end{table*}

\end{document}